\documentclass[twocolumn]{article}

\usepackage{amsfonts,amsmath}
\usepackage{authblk}
\usepackage[english]{babel}
\usepackage{booktabs}
\usepackage{float}
\usepackage{graphicx}
\usepackage{hyperref}
\usepackage[utf8]{inputenc}
\usepackage{interval}
\usepackage{natbib}
\usepackage{palatino} 
\usepackage{orcidlink}
\usepackage{siunitx}
\usepackage{subcaption}

\title{Photographic Visualization of Weather Forecasts with Generative Adversarial Networks}

\author{
Christian Sigg\thanks{Corresponding author: christian.sigg@meteoswiss.ch}\orcidlink{0000-0003-1067-9224}$^{1}$, Flavia Cavallaro\orcidlink{0000-0002-3699-0793}$^{2}$, Tobias Günther\orcidlink{0000-0002-3020-0930}$^{3}$ and Martin R. Oswald\orcidlink{0000-0002-1183-9958}$^{4,5}$\\
$^{1}$Federal Office of Meteorology and Climatology MeteoSwiss \qquad $^{2}$Comerge\\
$^{3}$Friedrich-Alexander-Universität Erlangen-Nürnberg\\
$^{4}$ETH Zürich \qquad $^{5}$University of Amsterdam
}

\date{}

\begin{document}

\maketitle

\begin{abstract}

Outdoor webcam images are an information-dense yet accessible visualization of past and present weather conditions, and are consulted by meteorologists and the general public alike. Weather forecasts, however, are still communicated as text, pictograms or charts. We therefore introduce a novel method that uses photographic images to also visualize future weather conditions.

This is challenging, because photographic visualizations of weather forecasts should look real, be free of obvious artifacts, and should match the predicted weather conditions. The transition from observation to forecast should be seamless, and there should be visual continuity between images for consecutive lead times. We use conditional Generative Adversarial Networks to synthesize such visualizations. The generator network, conditioned on the analysis and the  forecasting state of the numerical weather prediction (NWP) model,  transforms the present camera image into the future. The discriminator network judges whether a given image is the real image of the future, or whether it has been synthesized. Training the two networks against each other results in a visualization method that scores well on all four evaluation criteria.

We present results for three camera sites across Switzerland that differ in climatology and terrain. We show that users find it challenging to distinguish real from generated images, performing not much better than if they guessed randomly. The generated images match the atmospheric, ground and illumination conditions of the \mbox{COSMO-1} NWP model forecast in at least \SI{89}{\percent} of the examined cases. Nowcasting sequences of generated images achieve a seamless transition from observation to forecast and attain visual continuity.

\end{abstract}

\section{Introduction}
\label{sec:introduction}

\begin{figure*}
  \begin{subfigure}[t]{0.3\textwidth}
    \includegraphics[width=\textwidth]{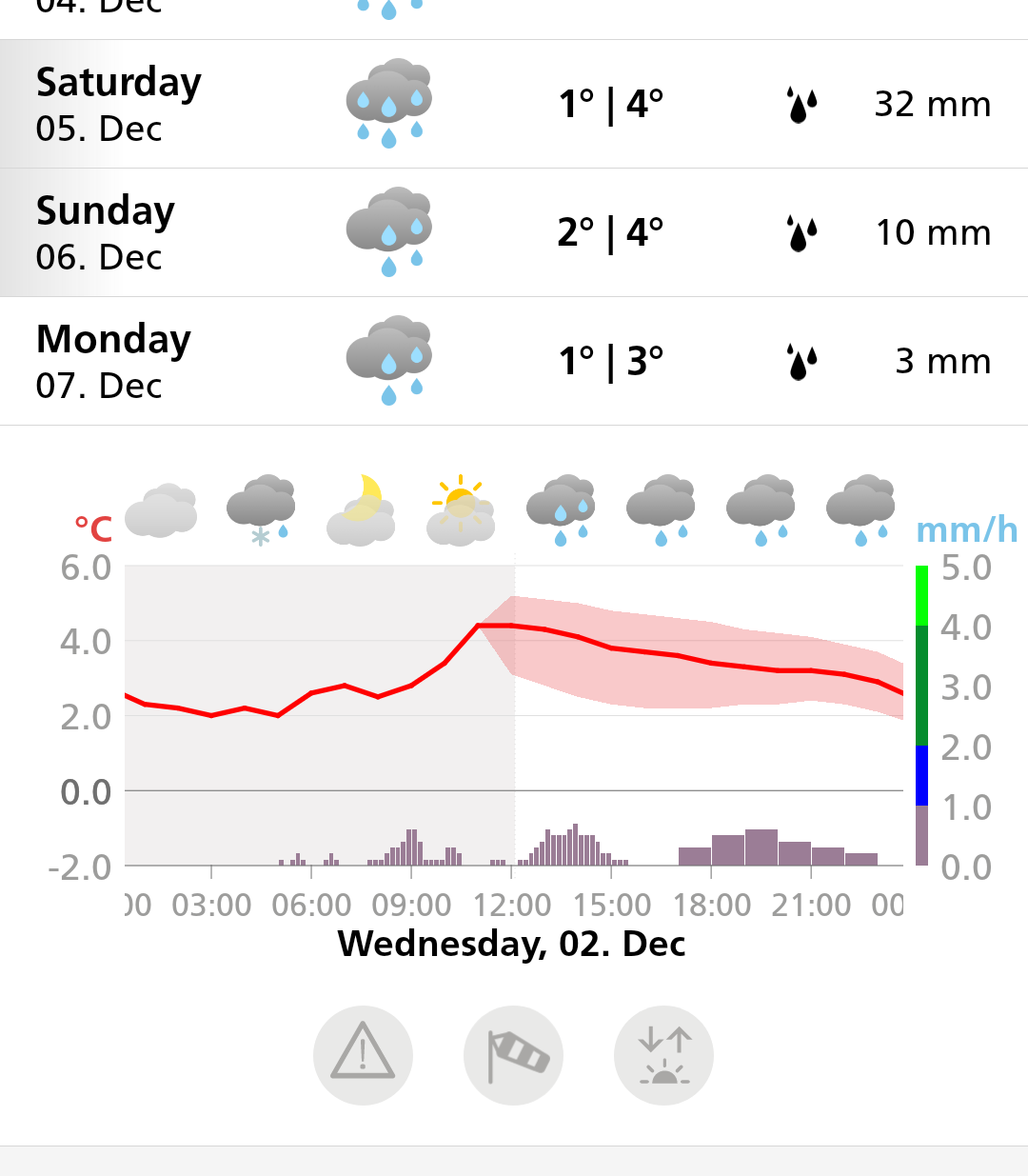}
    \caption{Air temperature and precipitation at a location of interest, visualized as line and bar charts. Pictograms provide additional summaries of cloud cover, sunshine and precipitation forecasts.}
    \label{fig:app_viz_chart}
  \end{subfigure}
  \hfill
  \begin{subfigure}[t]{0.3\textwidth}
    \includegraphics[width=\textwidth]{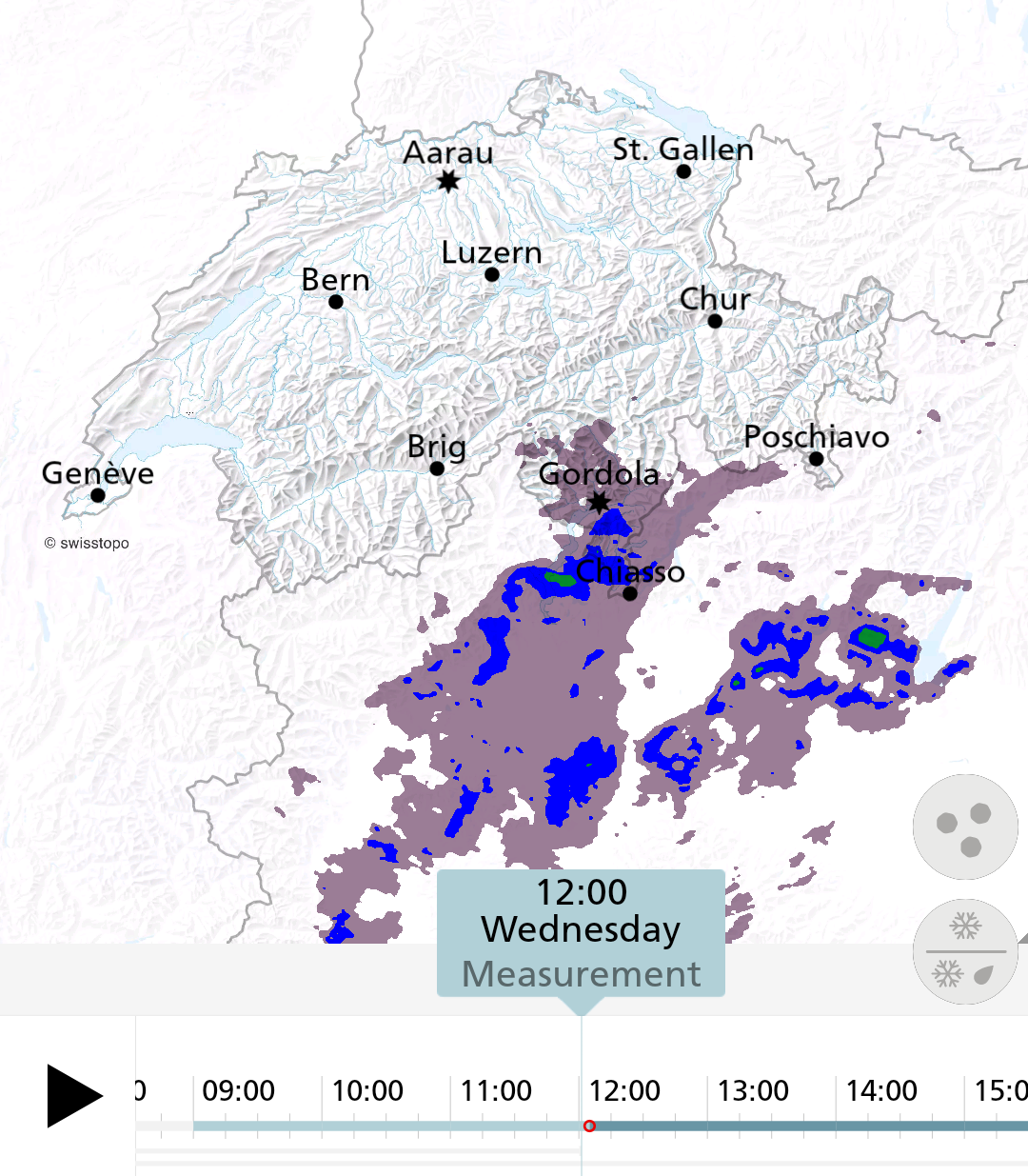}
    \caption{Animation of radar maps. By pushing the time slider into the future, the visualization provides a seamless transition from measurement to forecast.}
    \label{fig:app_viz_radar}
  \end{subfigure}
  \hfill
  \begin{subfigure}[t]{0.3\textwidth}
    \includegraphics[width=\textwidth]{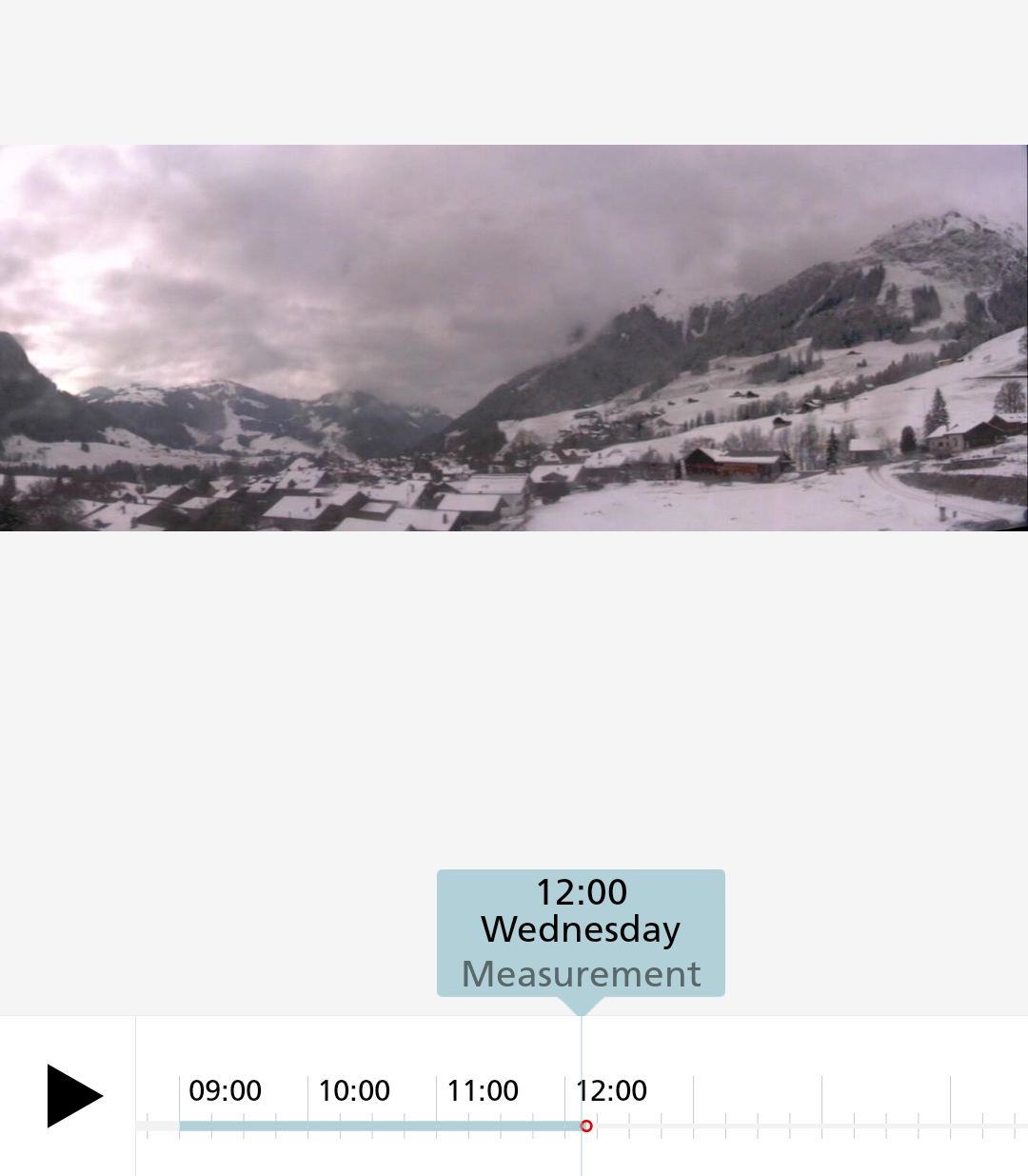}
    \caption{Animation of past and present images taken by a web camera. We envision an analogous transition from observed to synthesized images by pushing the time slider into the future.}
    \label{fig:app_viz_camera}
  \end{subfigure}
  \caption{Cropped screenshots of the MeteoSwiss smartphone app, showing different ways to seamlessly combine measurements and forecasts in the same visualization.}
\end{figure*}

Outdoor webcam images visualize past and present weather conditions, and are consulted by meteorologists and the general public alike, e.g.~in aviation weather nowcasting or the planning of a leisure outdoor activity. To this purpose, MeteoSwiss currently operates 40 cameras on measurement sites of its surface network, and provides the images to the public through its \href{https://www.meteoswiss.admin.ch/home/measurement-values.html?param=messnetz-webcams&table=true}{web page} and smartphone app (see Fig.~\ref{fig:app_viz_camera}). Rega (the Swiss air rescue service) operates a similar camera network for aviation weather forecasting and flight route planning. Private companies offer cameras as a service to communities and the tourism industry, operating hundreds of outdoor web cameras all across Switzerland.
However, these resources have not yet been utilized for forecasting.
Instead, weather forecasts are still communicated as text, pictograms or charts (Figures \ref{fig:app_viz_chart} and \ref{fig:app_viz_radar}). We  therefore introduce a novel visualization method that uses photographic images to show future weather conditions. As with the animation of radar maps (Fig.~\ref{fig:app_viz_radar}), we imagine a time slider that can be pushed beyond the present into the future (Fig.~\ref{fig:app_viz_camera}) to provide a seamless transition from observation to forecast.

Photographic visualizations of weather forecasts could have numerous applications. Meteorological services could use them to communicate localized forecasts over their own webcam feeds, smartphone apps and other distribution channels. They could also provide a service to communities and tourism organizations for creating forecast visualizations that are specific for their webcam feeds.

\subsection{Evaluation Criteria}

Forecast visualizations must satisfy several criteria to achieve their purpose. We propose the following four to evaluate the quality of photographic visualizations of weather forecasts:

\vspace{1 em}

\textbf{I. Realism.} The images should look real and be free of obvious artifacts. Ideally, it should not be possible to tell whether a given image was taken by an actual camera, or if it was synthesized by a visualization method.

\textbf{II. Matching future conditions.} The images should match the future atmospheric, ground and illumination conditions in the view of the camera. However, matching every pixel of a future observed image is not possible, as the forecast does not uniquely determine the positions and shapes of clouds, for example.

\textbf{III. Seamless transition.} The visualization method should achieve a seamless transition from observation to forecast. It should reproduce the present image, and must retain the present weather conditions as long as they persist into the future. 

\textbf{IV. Visual continuity.} The method should achieve visual continuity between images for consecutive lead times. For example, the movement of shadows should not show unnatural jumps across images.

\begin{figure*}
  \includegraphics[width=\textwidth]{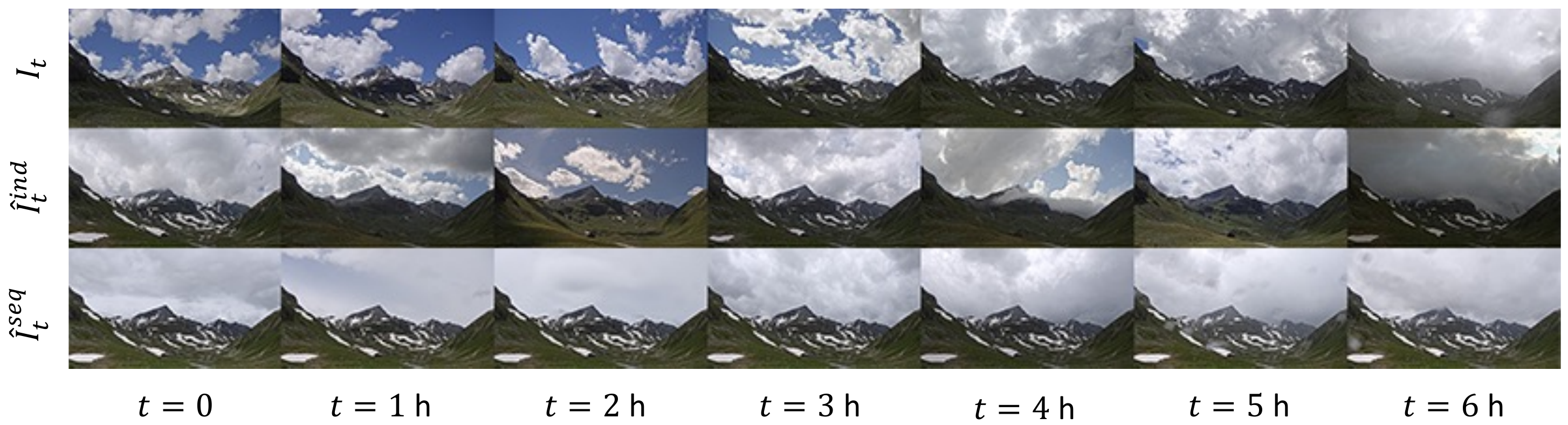}
  \caption{\emph{Top row:} A sequence of images $I_0, \dots, I_6$, taken by the Flüela camera in the Swiss Alps between 10:00 and 16:00 UTC on July 2nd, 2020. \emph{Middle row:} A visualization created using analog retrieval of individual images from an annotated archive (see Sec.~\ref{sec:data} for a description of the data). $\hat{I}^{ind}_t$ is the individual image from the archive where the associated weather descriptor $\hat{w}_t$ has the smallest Euclidean distance to the forecast $w_t$. \emph{Bottom row:} A visualization created using analog retrieval of image sequences. $\hat{I}^{seq}_0, \dots, \hat{I}^{seq}_6$ is the image sequence from the archive where the associated weather descriptors $(\hat{w}_0, \dots, \hat{w}_6)$ have the smallest Euclidean distance to $(w_0, \dots, w_6).$}
  \label{fig:analog_image_retrieval}
\end{figure*}

\subsection{Analog Retrieval}
\label{sec:analog_retrieval}

For a fixed camera view, such visualizations could be created using analog retrieval from an annotated image database. Given an archive of past images from the camera, annotated with the weather conditions that were present at the time the picture was taken, the forecast could be visualized by retrieving the images that most closely match the predicted conditions.

Analogs can be retrieved as individual images or whole sequences. Using \emph{per-image} analog retrieval, $\hat{I}^{ind}_t$ is the individual image from the archive where the associated weather descriptor $\hat{w}_t$ has the smallest Euclidean distance to the forecast $w_t$ for lead time $t$. As can be seen in the middle row of Fig.~\ref{fig:analog_image_retrieval}, using per-image analogs prioritizes matching the visible future weather conditions (our second evaluation criterion), but sacrifices the visual continuity between consecutive visualizations (our fourth criterion), with snow patches appearing and disappearing between images.

Using \emph{sequence} analog retrieval, $\hat{I}^{seq}_0, \dots, \hat{I}^{seq}_6$ is the image sequence where the associated weather descriptors $(\hat{w}_0, \dots, \hat{w}_6)$ have the smallest Euclidean distance to $(w_0, \dots, w_6)$. As can be seen in the bottom row of Fig.~\ref{fig:analog_image_retrieval}, this method satisfies the fourth criterion by construction. But to satisfy the second criterion would require that the archive contains whole sequences of images (and not just individual ones) that match the forecasted conditions.

Both methods trivially satisfy the first criterion, since the images were taken by the actual camera, but neither method achieves a seamless transition from observation to forecast (our third criterion) when $I_0$ shows distinctive atmospheric or ground conditions.

\subsection{Image Synthesis}

We therefore use image synthesis instead of image retrieval, which can be formalized as a regression problem in the following way. Given a description $w_t$ of the future weather conditions at time $t$, the \emph{generator} $G_0 \colon w_t\mapsto\hat{I}_t$ synthesizes an image $\hat{I}_t$. This image should closely match the real future image $I_t$, i.e. the dissimilarity measured by a suitable loss function $L(\hat{I}_t,I_t)$ should be small. Here, $G_0(w;\theta)$ is a neural network with parameters $\theta$, and is trained by minimizing the expected loss
\begin{equation}
\min_{\theta}\mathbb{E}_{w_t,I_t}\big[L\big(G_0(w_t;\theta),I_t\big)\big]
\label{eq:regression_objective}
\end{equation}
over pairs $(w_t, I_t)$ of weather forecasts and the corresponding real camera images. If the expected loss is small, then $G_0$ satisfies our second evaluation criterion.

\begin{figure}
  \includegraphics[width=\columnwidth]{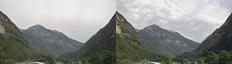}
  \caption{A pair of generated (\emph{left}) and real (\emph{right}) camera images, where $G_0$ was trained by minimizing the expected absolute difference of pixel values. While the ground and illumination conditions match quite well, the synthesized sky lacks detail, because it is not possible to predict the exact location and shape of the clouds from the weather forecast.}
  \label{fig:uniform_sky}
\end{figure}

The choice of $L$ is not obvious, however. Simple regression losses, such as the mean squared difference of pixel intensities, are not suitable for our application, since it is not possible to predict the exact location and shape of clouds from the weather forecast. Thus, a pixel-by-pixel equivalence of $\hat{I}_t$ and $I_t$ should not be sought. In fact, using a pixel-wise loss function results in images that show a mostly uniform sky (see Figure \ref{fig:uniform_sky}), unless $G_0$ is trained until it overfits and reproduces examples from the training data. Instead, $L$ should measure how well $\hat{I}_t$ matches the overall atmospheric, ground and illumination conditions of $I_t$. That is, a human examiner should not be able to tell whether $\hat{I}_t$ or $I_t$ is the real camera image of the future, even though they are not identical.

\subsection{Generative Adversarial Networks}
\label{sec:generative_adversarial_networks}

It is unclear how to design such a loss function by hand. \citet{goodfellow2014GenerativeAdversarialNets} introduced \textit{Generative Adversarial Networks} (GANs) as a method for learning the loss function from training data instead. Here, $G_1 \colon z\mapsto\hat{I}$ synthesizes an image from a random input $z$, which is sampled from a suitable distribution $z \sim p(z)$, for example the Gaussian distribution. A \emph{discriminator} $D_1 \colon I\mapsto \interval{0}{1}$ is introduced to mimic the human expert and to estimate the probability that the examined image $I$ is a real image, instead of having been synthesized by the generator. $D_1(I;\eta)$ is also a neural network, with parameters $\eta$. $G_1$ and $D_1$ are trained jointly and in an adversarial fashion by optimizing the objective
\begin{equation}
\min_{\theta}\max_{\eta}\mathbb{E}_{I}\!\big[\!\log D_1(I;\eta)\big]+\mathbb{E}_{z}\!\big[\!\log\! \big\{1-D_1\big(G_1(z;\theta);\eta\big)\!\big\}\!\big]
\label{eq:gan_objective}
\end{equation}
where $\mathbb{E}_{I}[\cdot]$ is the expected value w.r.t.~the distribution of real images. In this min-max optimization, the generator aims to fool the discriminator, and the discriminator tries to correctly distinguish between real and generated images. The training is complete when $G_1$ generates images with a high degree of realism and $D_1$ can no longer distinguish between real and generated images, thus satisfying our first evaluation criterion.

In order to synthesize images that not only look realistic, but are also corresponding to the weather forecast at lead time $t$, $w_t$ is provided as an additional input to both the generator and discriminator, $G_2(z|w_t; \theta)$ and $D_2(I|w_t;\eta)$. This \textit{conditional} adversarial training \citep{mirza2014ConditionalGenerativeAdversarial} ensures that the generated images match the weather forecast and satisfy our second evaluation criterion.

Our third goal is a seamless transition from the present image $I_0$ to generated future images $\hat{I}_t$. For $t=0$, the generator should therefore reproduce the present image, $\hat{I}_0=I_0$. This can be achieved by extending the generator and discriminator once more. Instead of synthesizing images from a random input $z$, $G_3(I_0, z | w_0, w_t)$  \textit{transforms} the current image $I_0$ into $\hat{I}_t$, based on the initial state $w_0$ and the forecast $w_t$ of the NWP model. $z$ adds a random component to the transformation, enabling the generator to synthesize more than one image that is consistent with the forecast when $t > 0$. The discriminator $D_3(I|I_0, w_0, w_t)$ is also conditioned on the full input. For the rest of the paper, we drop the subscripts and refer to $G_3$ and $D_3$ as $G$ and $D$.

GAN-based image transformation was introduced by \citet{isola2017ImagetoimageTranslationConditional}. The authors trained the networks using a linear combination of the regression objective in Eq.~\eqref{eq:regression_objective} and the adversarial objective in Eq.~\eqref{eq:gan_objective}. The $L_1$ norm was used as the pixel-wise regression loss, and the relative importance of the regression objective was set using a tuning parameter $\lambda$. We have found that in our application, setting $\lambda > 0$ sped up the training of $G$ for synthesizing the ground, but was detrimental for synthesizing clouds. Because their positions and shapes evolve over time, encouraging a pixel-wise consistency was again not helpful, and resulted in a sky lacking structure (as in the pure regression setting, Fig.~\ref{fig:uniform_sky}). We therefore only use the adversarial objective in Eq.~\eqref{eq:gan_objective} for training the generator and discriminator.

When the atmosphere is stable on the scale of the lead time step size, there will be only small changes from $I_t$ to $I_{t+1}$ in the position and shape of clouds. This should be reflected in the sequence of generated images $\hat{I}_0, \hat{I}_1, \hat{I}_2, \dots$ The results in Sec.~\ref{sec:visual_continuity} show that the generated sequences have a good degree of temporal continuity (the fourth criterion), even though we do not explicitly model the statistical dependency between consecutive lead times $t$ and $t+1$. Recurrent GANs \citep{mogren2016CRNNGANContinuousRecurrent} or adversarial transformers \citep{wu2020adversarial} are two potential avenues for further work in this area.

$G$ and $D$ can be trained for specific or arbitrary camera views. A view-independent generator could enable novel interactive applications, such as providing on-demand forecast visualizations for users equipped with a smartphone camera. But a view-independent generator will show its limits when the present view of the scenery is partially or fully blocked by opaque clouds or fog, and $w_t$ predicts a better visibility in the future. Although $G$ could generate natural looking scenery for the newly visible regions \citep{yu2018generative}, it will not be the real scenery of this location, thus potentially confusing the user. We therefore only consider the view-dependent case in this paper, where the synthesized and real scenery match very well (see Sec.~\ref{sec:matching_future_conditions}).

\section{Method}

\begin{figure*}
  \centering
  \includegraphics[width=0.9\textwidth]{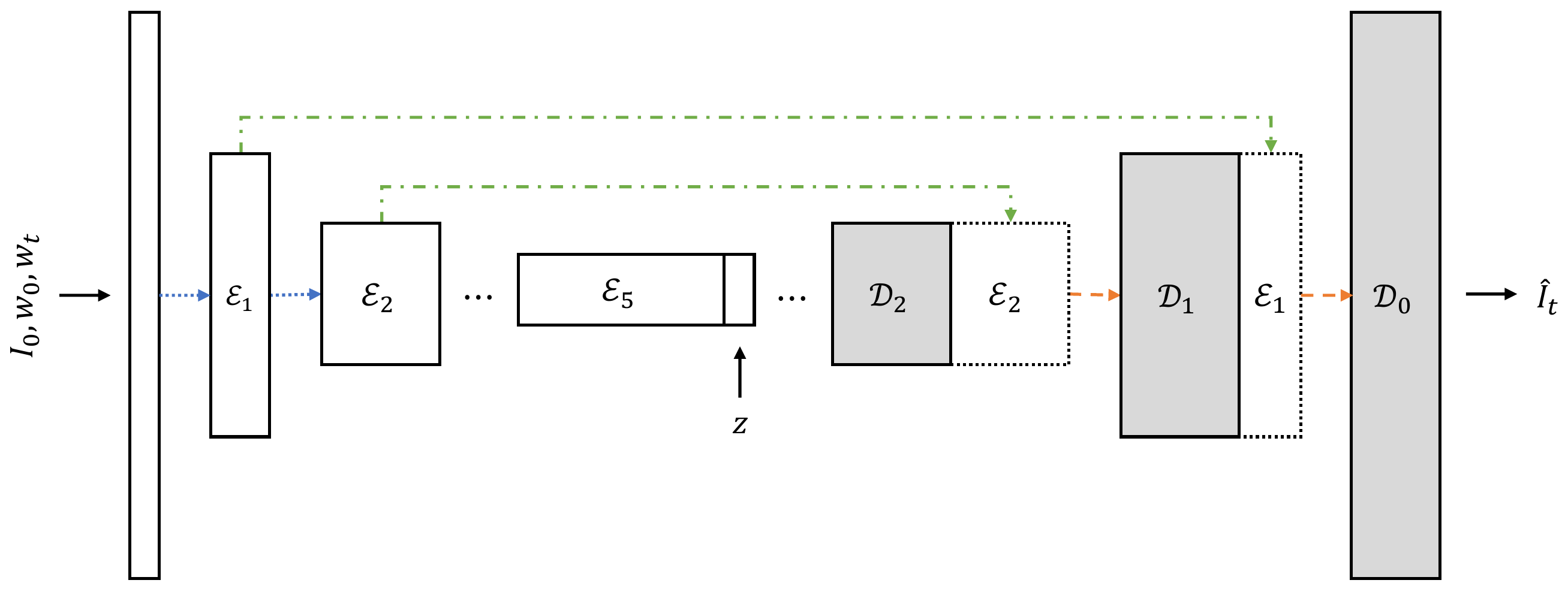}
  \caption{The encoder-decoder architecture of the generator network $G \colon I_0, w_0, w_t, z \mapsto \hat{I}_t$. Each encoder stage $\mathcal{E}_s$ halves the layer height and width and doubles the depth (blue dotted arrows), while each decoder stage $\mathcal{D}_s$ does the inverse (orange dashed arrows). The output of $\mathcal{E}_s$ is also concatenated to the output of the corresponding $\mathcal{D}_s$ (green dot-dashed arrows), providing additional long range connections that skip the in-between stages. A transformation of the random input $z \sim \mathcal{N}(0,1)$ is concatenated to the output of the innermost encoder stage $\mathcal{E}_5$. The complete architecture of $G$ has additional pre- and post-processing blocks and two more stages than shown in this figure.}
  \label{fig:architecture}
\end{figure*}

Neural networks have many design degrees of freedom, such as the number of layers, the number of neurons per layer, or whether to include skip connections between layers. The choice of the loss function and the optimization algorithm are also important, especially so for training GANs. Because the generator and discriminator are trained against each other, the optimization landscape is changing with every step. Possible training failure modes are a sudden divergence of the objective (sometimes after making progress for hundreds of thousands of optimization steps), or a collapse of diversity in the generator output \citep{goodfellow2014GenerativeAdversarialNets}. Past research therefore has focused on finding network architectures \citep[e.g.][]{radford2016UnsupervisedRepresentationLearning}, regularization \citep[e.g.][]{miyato2018SpectralNormalizationGenerative} and optimization schemes \citep[e.g.][]{heusel2017GANsTrainedTwo} that increase the likelihood of training success.

We present the network architecture, weight regularization and optimization algorithm that produced the results discussed in Sec.~\ref{sec:results}. We briefly also mention alternatives that we tried, but that did not lead to consistent improvements for our application.

\subsection{Network Architecture}
\label{sec:network_architecture}

Both $G$ and $D$ are encoder-decoder networks with skip connections (see Fig.~\ref{fig:architecture}), similar to a U-Net \citep{ronneberger2015UNetConvolutionalNetworks}. At each stage $\mathcal{E}_s$ of the encoder, the layer height and width is halved and the number of channels (the depth) is doubled, while the inverse is done at each decoder stage $\mathcal{D}_s$. By concatenating the output of $\mathcal{E}_s$ to the output of the corresponding $\mathcal{D}_s$, long-range connections are introduced that skip the in-between stages.

Each encoder stage $\mathcal{E}_s$ consists of three layers, a strided convolution (Conv) layer, followed by a batch normalization (BN) layer \citep{ioffe2015BatchNormalizationAccelerating} and a ReLU activation layer. The decoder stages have the same layer structure, except that a transposed convolution is used in the first layer.

A full specification of the networks is provided in the  \href{https://doi.org/10.5281/zenodo.6380148}{companion repository}. Here, we summarize the important architectural elements.

\textbf{Generator.} The input to $G$ consists of the three color channels of $I_0$ and the weather descriptors. The latter are concatenated as $w = \left(w_0, w_t\right)$, and repeated and reshaped to form additional input channels, such that each channel is the constant value of an element of $w$. After a Conv-BN-ReLU preprocessing block, there are five encoder stages $\mathcal{E}_1, \mathcal{E}_2, \dots, \mathcal{E}_5$ that progressively halve the input height and width and double the depth, except for the last stage, where the depth is not doubled anymore to conserve GPU memory.

The random input $z$ is sampled from a 100 dimensional Gaussian distribution with zero mean and unit variance. A dense linear layer and a reshaping layer transform $z$ into a tensor with the same height and width as the innermost encoder stage and a depth of 128 channels.

The input to the innermost decoder stage $\mathcal{D}_4$ consists of the output of the last encoder stage $\mathcal{E}_5$, concatenated with the transformed random input. Five decoder stages $\mathcal{D}_4, \mathcal{D}_3, \dots, \mathcal{D}_0$ restore the original height and width, followed by a Conv-BN-ReLU post-processing block and a final 1x1 convolution with a hyperbolic tangent activation that regenerates the three color channels.

\textbf{Discriminator.} The input to $D$ consists of the color channels of $I_0$ and $I_t$ and the weather descriptors $w_0$ and $w_t$, which are transformed into additional input channels as in $G$. The encoder and decoder again have five stages each, and there is a pre-processing block before the first encoding stage and a post-processing block after the last decoding stage. 

The output of the discriminator has two heads to discriminate between real and generated images on the patch and on the pixel level \citep{schonfeld2020UNetBasedDiscriminator}. The patch level output $D_{p}$ (see Sec.~\ref{sec:training_objective_and_optimizer}) is computed by combining the output channels of the last encoder stage with a 1x1 convolution. The pixel level output $D_{ij}$ is computed by a 1x1 convolution of the post-processing output.

\begin{figure}
  \includegraphics[width=\columnwidth]{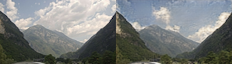}
  \caption{An example of visible artifacts introduced by a generator architecture that is based on residual blocks. Clouds in the input image (\emph{left}) are still partially visible in the clear sky regions of the output image (\emph{right}), as the residual transformation learned by the generator does not fully cancel their appearance.}
  \label{fig:residual_artifacts}
\end{figure}

 We also tried alternative network architectures based on residual blocks \citep[which add the block input to its output, see][]{he2015DeepResidualLearning}, similar to the BigGAN architecture \citep{brock2019LargeScaleGAN}. While the generator quickly learned to transform the ground, it created visible artifacts when transforming cloudy skies, see Fig.~\ref{fig:residual_artifacts} for an example. We conjecture that residual blocks are less suited for transformations of objects that move or evolve their shape over time, because the generator has to learn a transformation that perfectly cancels them from the block input if they are not to appear in the output.
 
\begin{figure}
  \includegraphics[width=\columnwidth]{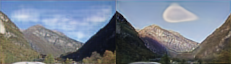}
  \caption{Examples of artifacts introduced by the upsampling method used in the generator. Nearest-neighbor upsampling often produced axis-aligned cloud patterns (\emph{left}), while bilinear upsampling produced blob-like cloud shapes with a smooth boundary (\emph{right}). Transposed convolutions avoided both problems.}
  \label{fig:upsampling_artifacts}
\end{figure}

We also tried replacing the strided convolution layers with nearest-neighbor or bilinear interpolation followed by regular convolution, which was suggested by \citet{odena2016deconvolution} to avoid checkerboard artifacts. This kind of artifact was not noticeable in the output of our final generator architecture, while nearest-neighbor upsampling often produced axis-aligned cloud patterns, and bilinear upsampling often produced overly smooth clouds, see Fig.~\ref{fig:upsampling_artifacts}.
 
\vspace{1 em}

Adding a second Conv-BN-ReLU block to each $\mathcal{E}_s$ and $\mathcal{D}_s$ also did not lead to a significant improvement, while doubling the number of trainable network weights.

\subsection{Training Objective and Optimizer}
\label{sec:training_objective_and_optimizer}

The training objective for $G$ and $D$ is an extension of Eq.~\eqref{eq:gan_objective}. We omit the dependency on the trainable network weights $\theta$ and $\eta$ in the following equations for the sake of brevity.

The discriminator objective to be maximized consists of a sum of three components. The first two components measure how well the patch-level head $D_p$ of the discriminator can distinguish between real
\begin{equation}
  \mathbb{E}_{I_{0},w_{0},I_{t},w_{t}}\left[\sum_{p}\log D_{p}(I_{t}|I_{0},w_{0},w_{t})\right]
  \label{eq:loss_disc_patch_real}
\end{equation}
and generated images
\begin{equation}
  \mathbb{E}_{I_{0},w_{0},w_{t}}\mathbb{E}_{z}\left[\sum_{p}\log\big[1-D_{p}\big(G(I_{0},z|w_{0,}w_{t})|\dots\big)\big]\right]
  \label{eq:loss_disc_patch_fake}
\end{equation}
The third component measures how well the pixel-level head $D_{ij}$ can distinguish between the real and generated pixels of a random \emph{cut-mix} composite $C$ \citep{yun2019CutMixRegularizationStrategy}
\begin{equation}
  \mathbb{E}_{C}\left[\sum_{i,\,j}M_{ij}D_{ij}(C)+\left(1-M_{ij}\right)\log\big(1-D_{ij}(C)\big)\right]
  \label{eq:loss_disc_pixel}
\end{equation}
The cut-mix operator combines a real and a generated image into a composite image $C$ using a randomly generated pixel mask $M$, see Fig. 3 in \citet{schonfeld2020UNetBasedDiscriminator} for an illustration, where $M_{ij}=0$ if the pixel $C_{ij}$ comes from the generated image, and $M_{ij}=1$ otherwise. Cut-mixing augments the training data, and $M_{ij}$ serves as the target label for the pixel-level head $D_{ij}$ of the discriminator. For our application, we apply the cut-mix operator to all the input channels of the discriminator, including the channels of the weather descriptors.

The generator objective to be minimized also consists of three components. The first two components measure how much the generator struggles to fool the discriminator on the patch level
\begin{equation}
  \mathbb{E}_{I_{0},w_{0},w_{t}}\mathbb{E}_{z}\left[\sum_{p}\log\big[D_{p}\big(G(I_{0},z|w_{0,}w_{t})|\dots\big)\big]\right]
  \label{eq:loss_gen_patch}
\end{equation}
and the pixel level
\begin{equation}
  \mathbb{E}_{I_{0},w_{0},w_{t}}\mathbb{E}_{z}\left[\sum_{i,\,j}\log\big[D_{ij}\big(G(I_{0},z|w_{0,}w_{t})|\dots\big)\big]\right]
  \label{eq:loss_gen_pixel}
\end{equation}
The third component measures how similar two generated images look at the pixel level, given different random inputs $z_1,z_2 \sim \mathcal{N}(0,1)$
\begin{equation}
  -\lambda\cdot\mathbb{E}_{I_{0},w_{0},w_{t}}\!\!\left[\!\sum_{i,\,j,\,c}\Big|G_{ijc}(I_{0},z_1|\dots) - G_{ijc}(I_{0},z_2|\dots)\Big|\!\right]
  \label{eq:loss_gen_similarity}
\end{equation}
where $G_{ijc}$ is the intensity of channel $c$ at pixel location $(i,j)$. Including this component in the objective for the generator avoids the problem of \emph{mode collapse}, where the generator ignores $z$ and produces a deterministic output given $(I_0,w_o,w_t)$. It also encourages the generator to make use of all stages of the encoder-decoder architecture, since $z$ is injected at the innermost stage. Setting $\lambda=1$ was sufficient to achieve both goals.

We evaluated three different optimization algorithms to train $\theta$ and $\eta$, stochastic gradient descent, \mbox{RMSprop} \citep{hinton2012neural} and ADAM \citep{kingma2017AdamMethodStochastic}. We have found that ADAM achieves the fastest loss reduction rate for our application, but that it can also produce erratic spikes in the loss curve. Spectral normalization of the training weights (see Sec.~\ref{sec:spectral_normalization}) and small learning rates were necessary to achieve a smooth training progress.

We also evaluated multiple discriminator updates per generator update, and different learning rates for the training of $G$ and $D$ \citep{heusel2017GANsTrainedTwo}. We found that using a two-times faster learning rate for the discriminator achieved the best results, and used \num{5e-5} as the learning rate for the discriminator and \num{1e-4} for the generator. The other ADAM hyperparameters were set to $\beta_1=0$ and $\beta_2=0.9$.

\subsection{Spectral Normalization}
\label{sec:spectral_normalization}

GANs are difficult to train, because the optimization landscape of the adversarial training changes with every iteration. The objective can spike suddenly (after many iterations of consistent improvement), or the training can stall completely if the discriminator becomes too good at distinguishing real from generated images.

We use spectral normalization of trainable weights \citep{miyato2018SpectralNormalizationGenerative} in all convolution layers to enforce Lipschitz continuity of the discriminator. As in \citet{zhang2019SelfAttentionGenerativeAdversarial}, we found that using spectral normalization also in the generator further stabilizes the training.

\section{Data}
\label{sec:data}

\begin{table*}
  \caption{The weather descriptor $w$ consists of the time of day, day of year, and the following subset of COSMO-1 output fields \citep[see][]{schattler2021COSMOModelVersion00}:}
  \label{tab:cosmo_fields}
  
  \begin{tabular}{lSl}
  \toprule
  Abbreviation & Unit & Name\tabularnewline
  \midrule
ALB\_RAD & \si{\percent} & Surface albedo for visible range, diffuse\tabularnewline
ASOB\_S & \si{W/m^2} & Net short-wave radiation flux at surface\tabularnewline
ASWDIFD\_S & \si{W/m^2} & Diffuse downward short-wave radiation at the surface\tabularnewline
ASWDIFU\_S & \si{W/m^2} & Diffuse upward short-wave radiation at the surface\tabularnewline
ASWDIR\_S & \si{W/m^2} & Direct downward short-wave radiation at the surface\tabularnewline
ATHB\_S & \si{W/m^2} & Net long-wave radiation flux at surface\tabularnewline
CLCH & \si{\percent} & Cloud area fraction in high troposphere (pressure below ca.~\SI{400}{hPa})\tabularnewline
CLCM & \si{\percent} & Cloud area fraction in medium troposphere (between ca.~\SIrange[range-phrase=~and~, range-units=single]{400}{800}{hPa})\tabularnewline
CLCL & \si{\percent} & Cloud area fraction in low troposphere (pressure above ca.~\SI{800}{hPa})\tabularnewline
CLCT & \si{\percent} & Total cloud area fraction\tabularnewline
D\_TD\_2M & \si{K} & \SI{2}{m} dew point depression\tabularnewline
DD\_10M & \si{\degree} & \SI{10}{m} wind direction\tabularnewline
DURSUN & \si{s} & Duration of sunshine\tabularnewline
FF\_10M & \si{m/s} & \SI{10}{m} wind speed\tabularnewline
GLOB & \si{W/m^2} & Downward shortwave radiation flux at surface\tabularnewline
H\_SNOW & \si{m} & Snow depth\tabularnewline
HPBL & \si{m} & Height of the planetary boundary layer\tabularnewline
PS & \si{Pa} & Surface pressure (not reduced)\tabularnewline
RELHUM\_2M & \si{\percent} & \SI{2}{m} relative humidity (with respect to water)\tabularnewline
T\_2M & \si{K} & \SI{2}{m} air temperature\tabularnewline
TD\_2M & \si{K} & \SI{2}{m} dew point temperature\tabularnewline
TOT\_PREC & \si{kg/m^2} & Total precipitation\tabularnewline
TOT\_RAIN & \si{kg/m^2} & Total precipitation in rain\tabularnewline
TOT\_SNOW & \si{kg/m^2} & Total precipitation in snow\tabularnewline
U\_10M & \si{m/s} & \SI{10}{m} grid eastward wind\tabularnewline
V\_10M & \si{m/s} & \SI{10}{m} grid northward wind\tabularnewline
VMAX\_10M & \si{m/s} & Maximum \SI{10}{m} wind speed\tabularnewline
  \bottomrule
  \end{tabular}
\end{table*}

We present results for three cameras. They belong to the networks operated by MeteoSwiss and Rega, where we have access to an archive of past images. We chose sites that show a good diversity in terms of terrain and weather conditions. The camera at Cevio (e.g. Fig.~\ref{fig:uniform_sky}) is located in the Maggia valley, in the southern part of Switzerland at an elevation of 421 m. The camera at Etziken (see Fig.~\ref{fig:nowcasting_illumination}) is located on the Swiss main plateau, at an elevation of 524 m. The camera at Flüela (see Fig.~\ref{fig:realism}) is located on a mountain pass in the Alps, at an elevation of 2177 m. 

Each camera takes pictures every 10 minutes, resulting in 144 images per day. We excluded images from our training and testing data sets that are not at all usable, such as when the camera moving head was stuck pointing to the ground, or the lens was completely covered with ice. But there was otherwise no need to  clean the data set. Images were retained if there were water droplets on the lens, or if there was a minor misalignment of the camera moving head. We also did not exclude particular weather conditions such as full fog. Nevertheless, there are gaps of several days in the data series, due to camera failures that could not be fixed in a timely manner.

To limit the memory consumption and to speed up the training on the available GPU (26 GB on a single NVIDIA A100), the images were downscaled from their original size to 64 by 128 pixels. Choosing powers of two for the height and width facilitates the down- and upsampling in the U-Net architecture, see Sec.~\ref{sec:network_architecture}. The original aspect ratios were restored using an additional horizontal resizing operation.

The weather descriptor $w$ consists of the time of day, day of year and a subset of the hourly forecast fields provided by the \mbox{COSMO-1} model (see \citet{schattler2021COSMOModelVersion00} and Table \ref{tab:cosmo_fields}), evaluated at the location of the camera. For the training and evaluation of the networks, we used the analysis (the initial state of the NWP model) as the optimal forecast for $w_0$ and also for $w_t$. This simplifies the evaluation of our visualization method, as it reduces the error between the NWP model state and the actual weather conditions at time $t$. For an operational use of the trained generator, where the analysis of the future is of course not available, one would use the current forecast for lead time $t$ instead, without having to modify the visualization method in any other way.

$w_0$ and $w_t$ can differ from the actual weather conditions visible in the camera image. The COSMO-1 fields have a limited spatial and temporal resolution of one kilometer and one hour. The camera location can thus be atypical for the NWP model grid point, and changes in weather conditions can be visible at the 10 min resolution of the camera before or after they occur in the model parameters. Using a single grid point can also be insufficient to capture the weather conditions in the view of the camera. When comparing the weather conditions visible in $I_t$ and $\hat{I}_t$ (see Sec.~\ref{sec:matching_future_conditions}), we therefore have to distinguish between mismatches that are due to the weather descriptors (where $w_0$ or $w_t$ do not accurately describe $I_0$ or $I_t$), and mismatches that are due to the generator (where $\hat{I}_t$ does not properly visualize $w_t$).

We used data from the year 2019 for training, and data from the year 2020 up to the end of August (when COSMO-1 was decommissioned at MeteoSwiss) for testing. The training data sets consist of all possible tuples $\left(w_0, w_t, I_0, I_t\right)$, where $t$ varies from zero to six hours. The resulting number of tuples are \num{712889} for Cevio, \num{450486} for Etziken and \num{493096} for Flüela.

\section{Results}
\label{sec:results}

We evaluate the forecast visualization method according to the four criteria that have been introduced in Sec.~\ref{sec:introduction}: how realistic the generated images look, how well they match future observed weather conditions, whether the transition from observation to forecast is seamless and whether there is visual continuity between consecutive visualizations.

Visualization methods transform data into a form that should be meaningful to humans. An evaluation of the four criteria is therefore best done using a visual inspection of the generated images. While there are objective measures for GANs that show some correlation to human judgment, such as the Fréchet Inception Distance \citep{heusel2017GANsTrainedTwo}, these measures have been developed and evaluated on image data sets that are very different from ours. We therefore used objective measures only to monitor the GAN training progress, but not in the final evaluation of the generated images.

\subsection{Realism}
\label{sec:realism}

\begin{table*}
  \caption{Results for the perceptual evaluation of real and generated images for the Cevio (left), Etziken (middle) and Flüela cameras (right). Using a browser-based data labeling tool, images were presented to five examiners with the question "What is your first impression of this image?", and they could answer either with "looks realistic" or "looks artificially generated". Their answers were aggregated into a confusion matrix for each camera, where the rows correspond to the ground truth, and the columns correspond to the examiners' judgment. The overall accuracy of the examiners was \SI{59}{\percent} for the Cevio camera, \SI{63}{\percent} for the Etziken camera and \SI{55}{\percent} for the Flüela camera.}
  \label{tab:realism}

  \begin{tabular}{rcc}
    \toprule
       & \multicolumn{2}{c}{Judgment} \tabularnewline
    \cmidrule(lr){2-3}
    Actual & Real & Generated \tabularnewline
    \midrule
    Real & 57 & 18 \tabularnewline
    Generated & 43 & 32 \tabularnewline
    \bottomrule
  \end{tabular}
  \hfill
  \begin{tabular}{rcc}
    \toprule
       & \multicolumn{2}{c}{Judgment} \tabularnewline
    \cmidrule(lr){2-3}
    Actual & Real & Generated \tabularnewline
    \midrule
    Real & 52 & 23 \tabularnewline
    Generated & 32 & 43 \tabularnewline
    \bottomrule
  \end{tabular}
  \hfill
  \begin{tabular}{rcc}
    \toprule
       & \multicolumn{2}{c}{Judgment} \tabularnewline
    \cmidrule(lr){2-3}
    Actual & Real & Generated \tabularnewline
    \midrule
    Real & 57 & 18 \tabularnewline
    Generated & 49 & 26 \tabularnewline
    \bottomrule
  \end{tabular}
\end{table*}

To evaluate the realism of individual generated images, we asked five coworkers at MeteoSwiss (who regularly consult the cameras of the MeteoSwiss network for their job duties, but otherwise were not involved in this project) to examine whether a presented image looks realistic or artificially generated.

The evaluation data was generated in the following way. For every camera, 75 pairs $(I_0, w_0)$ were sampled from the test data uniformly at random, but restricted to the hours between 06:00 and 14:00 UTC to avoid too many similarly looking nighttime views. For every pair, a lead time $t$ was sampled between \SIrange[range-phrase=~and~, range-units=single]{0}{360}{\minute} (in \SI{10}{\minute} increments), and both the real future image $I_t$ and the generated image $\hat{I}_t=G(I_0, z | w_0, w_t)$ were added to the evaluation data set. This sampling strategy ensured that there was no overall difference in the meteorological conditions of the real and generated images, which otherwise could have influenced the examiners' accuracy.

\vspace{2 em}

Each examiner was then assigned 30 randomly selected images from each camera, and asked to provide their judgment on the realism of each presented image. The examiners could inspect the images at arbitrary zoom levels and look at other images in the evaluation set before giving an answer. But they were not given any background information about the images, such as the date, the lead time of the forecast or the predicted weather conditions.

\begin{figure}
  \includegraphics[width=\columnwidth]{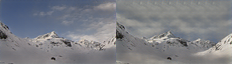}
  \caption{Examples of generated images that were judged to look realistic (\emph{left}), and that contain obvious artifacts such as repeating cloud patterns (\emph{right}).}
  \label{fig:realism}
\end{figure}

The results of the evaluation are presented in Table~\ref{tab:realism}. The overall low accuracy of the examiners' judgment indicates that it was challenging for them to distinguish between real and generated images (a \SI{50}{\percent} accuracy would correspond to random guessing). Many of the generated images look realistic enough to pass for a real image, but there were also instances where artifacts introduced by the generator were obvious at first glance (see Fig.~\ref{fig:realism} for an example).

\begin{figure*}
  \includegraphics[width=\textwidth]{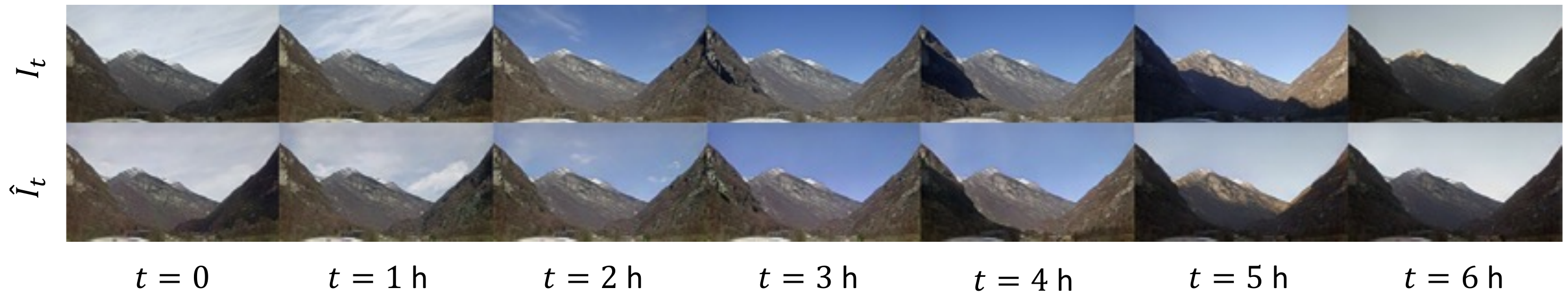}
  \caption{A sequence of images taken by the Cevio camera between 10:00 and 16:00 UTC on February 6th, 2020 (\emph{top row}), and the corresponding images generated by $G(I_0, z | w_0, w_t)$ (\emph{bottom row}). This visualization satisfies our four evaluation criteria. The generated images look realistic and are free of artifacts. They match the real images of the future w.r.t.~atmospheric, ground and illumination conditions. The transition from observation to forecast is seamless: $I_0$ (\emph{top left}) is well approximated by $\hat{I}_0$ (\emph{bottom left}), and $\hat{I}_1$ retains the persisting weather conditions of $I_0$. Finally, the progression of shadows appears natural.}
  \label{fig:nowcasting_overall}
\end{figure*}

\subsection{Matching Future Conditions}
\label{sec:matching_future_conditions}

To evaluate how well the forecast visualization $\hat{I}_t$ matches the future real image $I_t$, we compared the  atmospheric, ground and illumination conditions visible in the images. As we do not expect $\hat{I}_t$ to match $I_t$ in every detail (e.g.~the precise shape and position of clouds do not matter), we used the following descriptive criteria to determine their overall agreement.

\textbf{Atmospheric conditions.} Cloud cover: \textit{clear sky \textbar\space few \textbar\space cloudy \textbar\space overcast}, cloud type: \textit{cumuliform \textbar\space stratiform \textbar\space stratocumuliform \textbar\space cirriform} and  visibility: \textit{good \textbar\space poor}

\textbf{Ground conditions.} \textit{Dry \textbar\space wet \textbar\space frost \textbar\space snow}

\textbf{Illumination.} Time of day: \textit{dawn \textbar\space daylight \textbar\space dusk \textbar\space night}, sunlight: \textit{diffuse \textbar\space direct (casting shadows)}

A mismatch between the actual and visualized conditions can happen because of two different kinds of failures. The forecast descriptor $w_t$ can fail to accurately capture the weather conditions visible in $I_t$, or the generated image $\hat{I}_t$ can be inconsistent with $w_t$:

\textbf{Inaccurate forecast.} Besides COSMO-1 not being a perfect forecast model, evaluating the forecast fields only at the camera site can be insufficient to describe all of the visible weather. Furthermore, the spatial or temporal resolution of $w_t$ can be too coarse for highly variable weather conditions. For example, the hourly granularity of the forecast can only approximately predict the time when it starts to rain.

\textbf{Inconsistent visualization.} The generator $G$ can fail to properly account for the changes from $w_0$ to $w_t$ in the transformation of $I_0$ to $\hat{I}_t$.

We used 45 pairs of the same data set as in Sec.~\ref{sec:realism}. For each pair $(I_t, \hat{I}_t)$, we counted the matching descriptive criteria (at most three for the atmospheric conditions, one for the ground condition and two for the illumination conditions). If there was a mismatch in a criterion, we checked whether $w_t$ did not accurately describe $I_t$ (the first kind of failure), or whether $\hat{I}_t$ was inconsistent with $w_t$ (the second kind of failure). Only if $w_t$ was accurate, but $\hat{I}_t$ was inconsistent with it, did we consider the visualization to have failed.

\begin{figure}
  \includegraphics[width=\columnwidth]{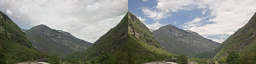}
  \caption{The cloud amount in the generated image (\emph{left}) is too large, compared to the observed conditions (\emph{right}) at Cevio on April 25th 2020 at 12:00 UTC. But the mismatch is not a failure of the visualization method, as the COSMO-1 forecast predicts a \SI{100}{\percent} cloud area fraction in the medium troposphere.}
  \label{fig:mismatching_visibility}
\end{figure}

Table \ref{tab:matching_conditions} summarizes the results of the evaluation. In general, cloud cover and cloud type were the most difficult criteria to get right. At Cevio for example, only in 32 out of 45 cases did the cloud cover match the observed conditions. But in only 5 cases was the mismatch due to $G$, i.e.~$w_t$ was accurate but $\hat{I}_t$ was inconsistent with it. The other times when there was a mismatch, $w_t$ did not accurately describe $I_t$ (see Fig.~\ref{fig:mismatching_visibility} for an example). The visualization method would therefore benefit the most from improving the accuracy of the weather forecast, e.g.~by using a post-processed version of the COSMO-1 forecast with a higher spatial and temporal resolution.

\begin{table*}
    \caption{Counts of matching atmospheric, ground and illumination conditions between real and generated images $I_t$ and $\hat{I}_t$ at time $t$. For each camera, 45 random pairs $(I_t, \hat{I}_t)$ were visually compared according to the criteria of Sec.~\ref{sec:matching_future_conditions}. Whenever there was a mismatch in at least one criterion, we checked if the weather descriptors $w_t$ of the forecast accurately describe the conditions visible in $I_t$. Only in this case did we count the visualization to have failed.}
  \label{tab:matching_conditions}
  
  \begin{tabular}{lccccccc}
    \toprule
      & \multicolumn{6}{c}{Matching conditions} & \tabularnewline
    \cmidrule(lr){2-7}
      & \multicolumn{3}{c}{Atmosphere} &  & \multicolumn{2}{c}{Illumination} & \tabularnewline
    \cmidrule(lr){2-4} \cmidrule(lr){6-7} 
    Camera & Cloud cover & Cloud type & Visibility & Ground & Time of day & Diffuse/direct & Viz.~failures\tabularnewline
    \midrule
    Cevio & 32 & 35 & 45 & 45 & 45 & 40 & 5\tabularnewline
    Etziken & 36 & 36 & 44 & 45 & 45 & 38 & 2\tabularnewline
    Flüela & 31 & 33 & 26 & 44 & 41 & 35 & 5\tabularnewline
    \bottomrule
  \end{tabular}
\end{table*}

Figures \ref{fig:nowcasting_overall} to \ref{fig:nowcasting_snowpatch} show further examples of the range of transformations that can be achieved by the generator, transforming the atmospheric and illumination conditions to match $I_t$, while retaining the specific ground conditions of $I_0$.

\subsection{Seamless Transition}
\label{sec:seamless_transition}

For a seamless transition between observation and forecast, the generator must be able to reproduce the input image $\hat{I}_0 = I_0$ for $t=0$. It must also retain the conditions of $I_0$ in $\hat{I}_1, \hat{I}_2, \dots$ as long as they persist into the future. 

We evaluated the third criterion on hourly nowcasts up to six hours into the future. As can be seen in Figures \ref{fig:nowcasting_overall} to \ref{fig:nowcasting_sun}, the generator reproduces the ground and illumination conditions of $I_0$ very well in $\hat{I}_0$. The shape and positions of clouds are reproduced closely but not exactly, giving the impression that the generator reconstructs the overall cloud pattern, but not every small detail. We could achieve pixel-level accurate reproductions of $I_0$ by using a residual architecture for the generator network. But as discussed in Sec.~\ref{sec:network_architecture}, the realism of images generated using a residual network suffered from noticeable visual artifacts for $t>0$.

As can be seen in the bottom row of Fig.~\ref{fig:nowcasting_snowpatch}, the generator retains the specific conditions of $I_0$ (such as the position and shape of snow patches) in the visualizations $\hat{I}_1, \dots, \hat{I}_6$. This is only possible because $G$ \emph{transforms} $I_0$ into $\hat{I}_t$. As already shown in the corresponding Fig.~\ref{fig:analog_image_retrieval}, a seamless transition cannot be achieved using analog retrieval (Sec.~\ref{sec:analog_retrieval}), because an image with the specific shapes and positions of clouds and snow patches will not exist in the archive.

\subsection{Visual Continuity}
\label{sec:visual_continuity}

\begin{figure*}
  \centering
  \includegraphics[width=0.75\textwidth]{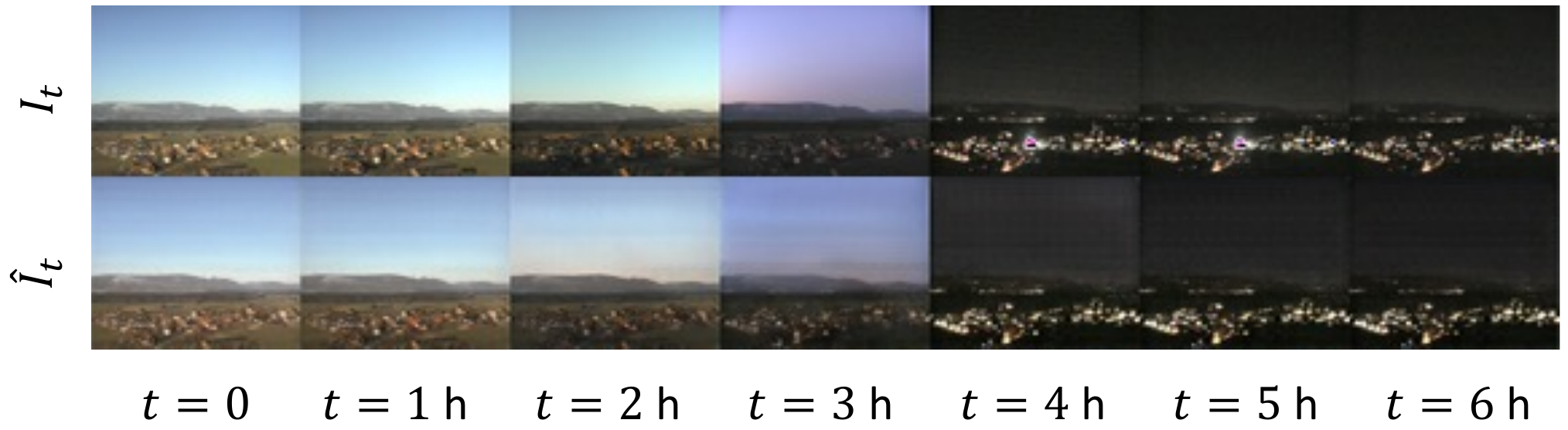}
  \caption{A sequence of images taken by the Etziken camera between 14:00 and 20:00 UTC on February 6th, 2020 (\emph{top row}), and the corresponding images generated by $G(I_0, z | w_0, w_t)$ (\emph{bottom row}). The generator was able to fully transform $I_0$ (\emph{top left}) from daylight to nighttime conditions, including the visual appearance of illuminated street lamps and windows at $t=\SI{4}{\hour}$.}
  \label{fig:nowcasting_illumination}
\end{figure*}

\begin{figure*}
  \includegraphics[width=\textwidth]{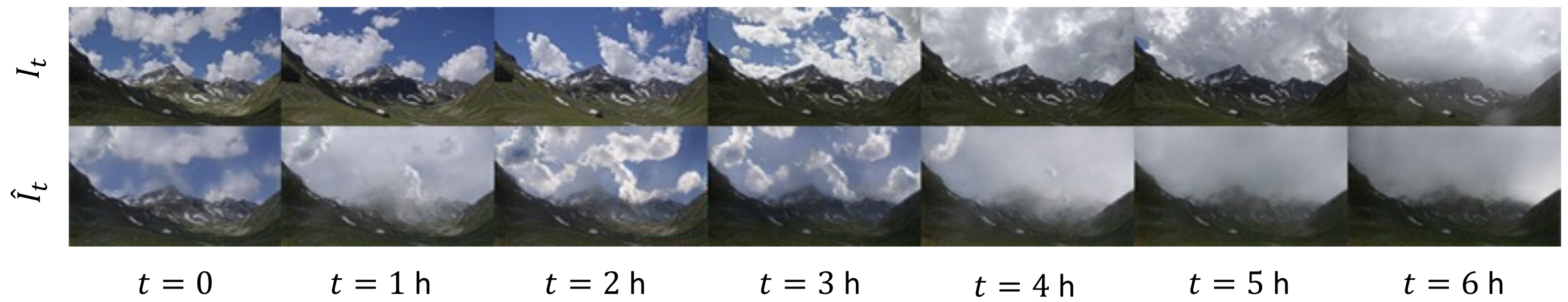}
  \caption{\emph{Top row:} The same sequence of images as in the top row of Fig.~\ref{fig:analog_image_retrieval}, taken by the Flüela camera between 10:00 and 16:00 UTC on July 2nd, 2020. \emph{Bottom row:} Because $\hat{I}_t=G(I_0, z | w_0, w_t)$  is a \emph{transformation} of $I_0$, the exact position and shape of snow patches are retained in $\hat{I}_1, \hat{I}_2, \dots, \hat{I}_6$.}
  \label{fig:nowcasting_snowpatch}
\end{figure*}

\begin{figure*}
  \includegraphics[width=\textwidth]{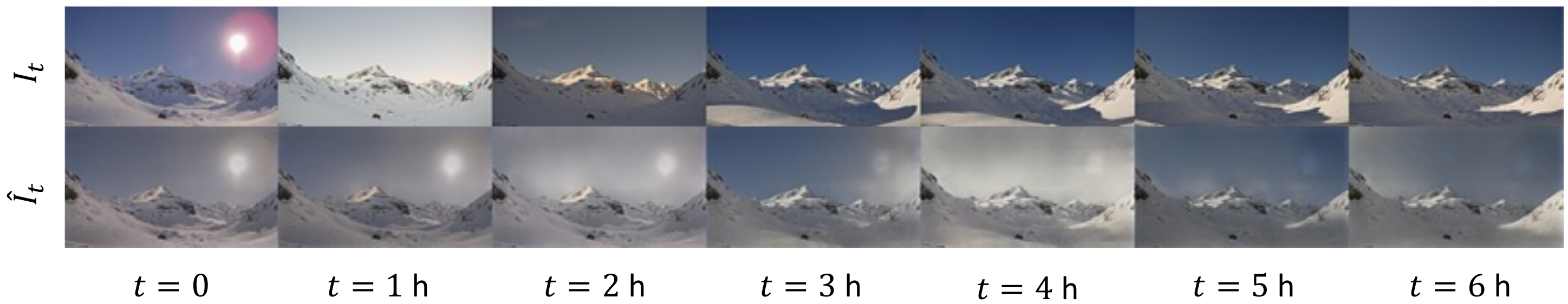}
  \caption{\emph{Top row:} A sequence of images taken by the Flüela camera between 06:00 and 12:00 UTC on January 12th, 2020. \emph{Bottom row:} The generator  accurately transformed the illumination conditions and shadows. But it failed to learn how the sun moves across the sky. Instead of shifting the position of the sun, the sun stays in the same position and fades away gradually.}
  \label{fig:nowcasting_sun}
\end{figure*}

\begin{figure*}
  \includegraphics[width=\textwidth]{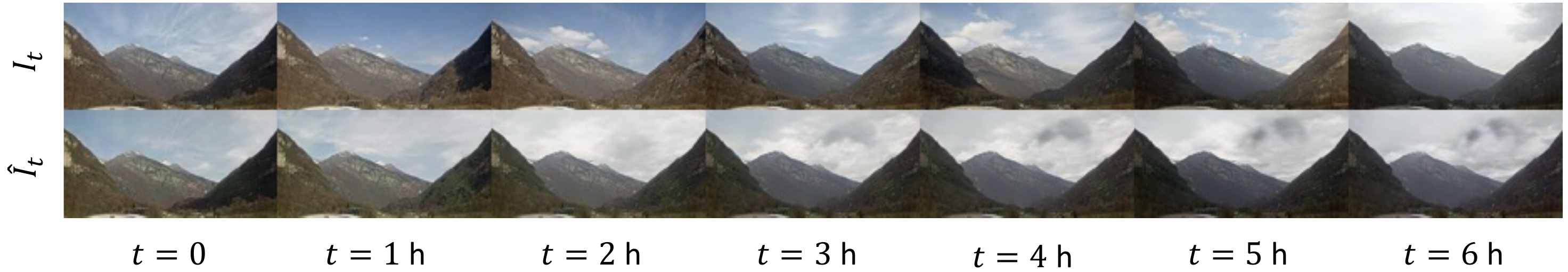}
  \caption{\emph{Top row:} A sequence of images taken by the Cevio camera between 10:00 and 16:00 UTC on February 17th, 2020.  \emph{Bottom row:} The generated images accurately match the ground and illumination conditions, and the development of cloud shapes appears natural. But the \emph{positions} of individual clouds remain too static over time.}
  \label{fig:nowcasting_static_clouds}
\end{figure*}

Finally, the consecutive visualizations $\hat{I}_t, \hat{I}_{t+1}$ must show visual continuity, with a natural looking cloud development, change of daylight, movement of shadows, and so on. As can be seen in Figures \ref{fig:nowcasting_overall} to \ref{fig:nowcasting_static_clouds}, the evolution of the ground and illumination conditions matches the future observed images very closely, even though $G$ and $D$ do not explicitly account for the statistical dependency between $t$ and $t+1$. The increase or decrease in cloud cover and visibility also looks natural. By contrast, the visualizations based on individual analogs $\hat{I}^{ind}_t$ lack continuity (Fig.~\ref{fig:analog_image_retrieval}, middle row), with snow patches appearing and disappearing, and unnatural changes in the illumination conditions.

But $G(I_0, z | w_0, w_t)$ struggled with learning image transformations that involve translations of objects across the camera view, such as the movement of the sun (Fig.~\ref{fig:nowcasting_sun}) or of isolated clouds (Fig.~\ref{fig:nowcasting_static_clouds}). We conjecture that a network architecture based on Conv-BN-ReLU blocks is highly effective at transforming the appearance of objects that remain in place, but less for translations. We therefore investigated whether including non-local network layers, such as self-attention \citep{zhang2019SelfAttentionGenerativeAdversarial}, could be beneficial. But we did not achieve a consistent improvement in our experiments, neither with full nor with axis-aligned self-attention, whereas the network training time increased significantly.

\begin{figure*}
  \includegraphics[width=\textwidth]{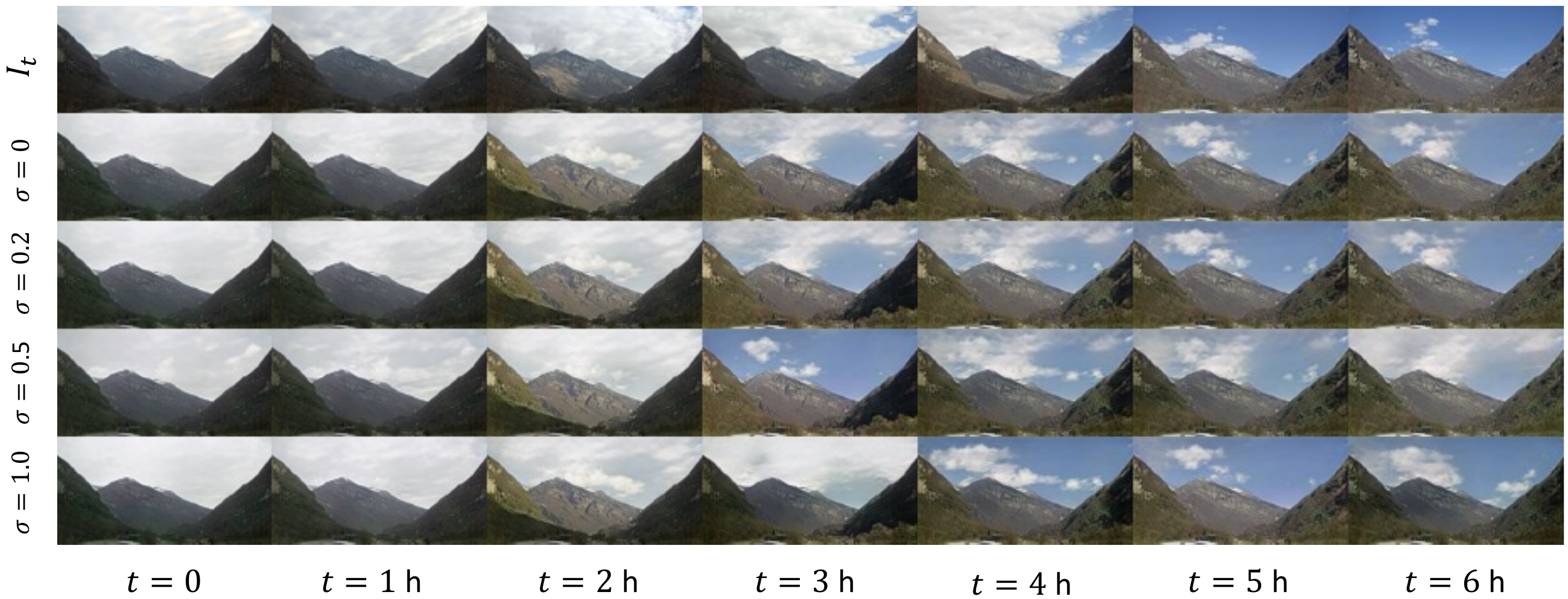}
  \caption{The effect of the noise variance on the diversity of visualizations synthesized by the generator. Increasing $\sigma$ when sampling $z \sim \mathcal{N}(0,\sigma^2)$ leads to a greater diversity of images that are deemed consistent with the weather forecast. But note that the realised diversity is also a function of $t$: it is smallest at $t=0$, and greatest when there is a change of weather conditions at $t=\SI{3}{\hour}$ . \emph{Top row:} A sequence of images taken by the Cevio camera between 6:00 and 12:00 UTC on March 16th, 2020. \emph{Following rows:} Nowcasting visualizations generated with $\sigma=(0,0.2,0.5,1.0)$.}
  \label{fig:nowcasting_sampling_variance}
\end{figure*}

The balance between visual continuity and image diversity can be tuned by changing the standard deviation $\sigma$ of the random input $z \sim \mathcal{N}(0,\sigma^2)$. Increasing $\sigma$ leads to a greater diversity of visualizations that are deemed consistent with the weather forecast, while decreasing $\sigma$ promotes greater continuity between subsequent images (see Fig.~\ref{fig:nowcasting_sampling_variance}). We have found that setting $\sigma=0.5$ results in the best trade-off between the two objectives.

\section{Conclusions and Future Work}

We have shown that photographic images can not only visualize the weather conditions of the past and the present, but can be useful for weather forecasts as well. Using conditional Generative Adversarial Networks, it is possible to synthesize photographic visualizations that look realistic, match the predicted weather conditions, transition seamlessly from observation to forecast, and show a high degree of visual continuity between consecutive forecasting lead times.

Meteorological services could use such visualizations to communicate their localized forecasts in an additional format that is immediately accessible to the user. They could also provide a service to communities and tourism organizations for creating forecast visualizations that are specific for their web camera feeds. 

The visualization method introduced in this paper is mature enough to become the first generation of an operational forecast product. But there are several next steps which could improve its visual fidelity and accuracy. 

Training GANs is computationally intensive, which is why we had to limit the image size to 64 by 128 pixels. But there exist techniques in the literature \citep[e.g.][]{karras2018ProgressiveGrowingGANs} that scale GANs to image sizes of at least a megapixel. Training the production grade high-resolution network could be done in the cloud on sufficiently capable hardware.

Our results of Sec.~\ref{sec:matching_future_conditions} indicate that the generator is rarely to blame for mismatches between the actual and the visualized conditions. Errors occur instead if the weather descriptor fails to accurately capture the conditions visible in the image. Using sub-hourly and sub-kilometer forecasts could improve the accuracy of the weather descriptors, and therefore the accuracy of the visualization method. Evaluating the forecast output fields at multiple locations in the line of sight of the camera could also be beneficial.

While the temporal evolution of the ground and illumination conditions looks natural, our GAN architecture still struggles with synthesizing the movement of the sun and isolated clouds across the sky. We will continue to investigate non-local network layers that could complement the convolution layers. We will also investigate whether synthesizing whole sequences of images (instead of single images) can further improve the temporal evolution of nowcast visualizations.

Finally, if it is possible to quickly adapt a view-independent generator to a specific view (ideally with a single image), novel interactive applications of photographic visualizations would also become possible. Smartphone users could obtain personalized forecasts by taking an image of the local scenery and a reading of their geographic coordinates, and obtain a photographic animation that shows the predicted weather conditions in their near future.

\section*{Acknowledgments}

We thank Rega for giving us the permission to use images from the Cevio camera in this study.

We thank Tanja Weusthoff for the preparation of the \mbox{COSMO-1} forecast data.

We thank Christian Allemann, Yannick Bernard, Eliane Thürig, Deborah van Geijtenbeek and Abbès Zerdouk for evaluating the realism of individual generated images.

We thank Daniele Nerini for providing his expertise on nowcasting and post-processing of forecasts.

\section*{Code and Data Availability}

A Tensorflow implementation of the generator and discriminator networks, as well as the code to reproduce the experiments presented in Sec.~\ref{sec:results}, is available in the  \href{https://doi.org/10.5281/zenodo.6374374}{companion repository}. 

The repository also contains the trained networks for the three camera locations, and additional generated images (of which Figures \ref{fig:nowcasting_overall} to \ref{fig:nowcasting_sampling_variance} are examples). The data used in the expert evaluations and their detailed results (summarized in Tables \ref{tab:realism} and \ref{tab:matching_conditions}) are also available there.

The complete image archive and \mbox{COSMO-1} data used for the training and evaluation of the networks cannot be published online due to licensing restrictions. But it can be obtained free of charge for academic research purposes by contacting the \href{https://www.meteoswiss.admin.ch/home/form/customer-service.html}{MeteoSwiss customer service desk}.

\bibliography{bibliography}
\bibliographystyle{plainnat}

\end{document}